\documentclass[runningheads]{llncs}

\usepackage{eccv}

\usepackage{eccvabbrv}
\usepackage{bbding}
\usepackage{graphicx}
\usepackage{booktabs}
\usepackage{float}
\usepackage{caption} 
\usepackage{multirow}
\usepackage[accsupp]{axessibility}  
\definecolor{myhl}{HTML}{E1F4FC}
\definecolor{green1}{HTML}{82B366}
\definecolor{red1}{HTML}{B85450}
\definecolor{mainbg}{HTML}{E1F4FC}
\definecolor{sidebg}{HTML}{fdc7cf}
\newcommand{\equalcontrib}{\textsuperscript{*}}
\newcommand{\correspondingauthor}{\textsuperscript{\textdagger}}

\usepackage{hyperref}
\usepackage{lipsum}
\usepackage{orcidlink}
\usepackage{pifont}
\usepackage[table]{xcolor}
\usepackage{wrapfig} 
\begin{document}

\title{Tri-Efficient Transfer Learning for Point Cloud Videos} 

\titlerunning{Tri-Efficient Transfer Learning for Point Cloud Videos}

\author{Yiding Sun\inst{1}\equalcontrib \and
Dongxu Zhang\inst{1}\equalcontrib \and
Jihua Zhu\inst{1}\correspondingauthor\and
Haozhe Cheng\inst{1}\and
Zhengqiao Li\inst{2}\and
Pengcheng Li\inst{3}\and
Chaowei Fang\inst{1}\and
Yonghao Dong\inst{1}\and
Lin Chen\inst{1}}

\authorrunning{Y. Sun et al.}

\institute{School of Software Engineering, Xi'an Jiaotong University \\ \email{sunyiding@stu.xjtu.edu.cn, zhujh@xjtu.edu.cn}
\and
School of Information and Communication Engineering, Xi'an Jiaotong University
\and
SIGS, Tsinghua University\\
\textsuperscript{*}Equal contribution.
\quad
\textsuperscript{\textdagger}Corresponding author.}
\maketitle

\begin{abstract}
While point cloud foundation models have significantly advanced point cloud video understanding, existing parameter-efficient fine-tuning (PEFT) methods still suffer from two critical limitations: prohibitive annotation costs for large-scale point cloud datasets and severe memory bottlenecks. In this paper, we aim to mine richer supervision signals from existing data rather than blindly scaling datasets. A further key principle is that the memory footprint of fine-tuning must be drastically reduced compared to full fine-tuning, which remains elusive for current PEFT techniques. Driven by these challenges, we identify three core desiderata: data-, parameter-, and memory efficiency, and present PoinTriE, a unified framework that excels along all three dimensions. For pre-training, pseudo-motion trajectories are synthesized via rigid transformations, paired with text corpora and 2D projections derived from raw point clouds. We then propose a Geometric-Motion Duality Network optimized via multimodal contrastive learning, rigid rotation prediction, and motion distribution divergence to produce dense self-supervision. During fine-tuning, we freeze the pretrained backbone and only update a lightweight Spatio-temporal Side Network built with LoRA units. Equipped with a gradient flow masking strategy, PoinTriE simultaneously reduces memory consumption and parameter overhead. Extensive experiments confirm that PoinTriE establishes new state-of-the-art results on action recognition and semantic segmentation tasks.
\end{abstract}

\section{Introduction}
\label{sec:intro}
Fine-tuning pretrained point cloud foundation models (PFMs) has been proven remarkably effective in adapting to versatile scenarios, especially those in dynamic domains~\cite{bao2025gaussian,bao2026distractor}. However, catastrophic forgetting and prior knowledge disruption occur frequently, particularly as model sizes continue to grow~\cite{qi2023contrast,li2025pointdico}. To overcome these challenges, Parameter-Efficient Fine-Tuning (PEFT) methods~\cite{zha2025pma,zhou2024dynamicdapt,wang2025pointlora}, such as PointCSA~\cite{csa}, have emerged as more promising alternatives, garnering significant attention. 

Instead of updating all model parameters directly, PointCSA inserts a learnable temporal adapter into the intertwined feature space of Transformer~\cite{vaswani2017attention} blocks, bridging the gap between static and dynamic domains while keeping the PFM frozen. This method significantly reduces the number of parameters requiring fine-tuning, thereby reducing computational overhead. Follow-up research has further enhanced this pipeline by improving aspects such as adapter architecture~\cite{zha2025pma,sun2026alignadaptrethinkingparameterefficient}, local geometric interaction~\cite{gstliang2025parameter}, and backbone model scale~\cite{deng2024vg4d}.

\begin{figure}[tb]
  \centering
  \includegraphics[width=\linewidth]{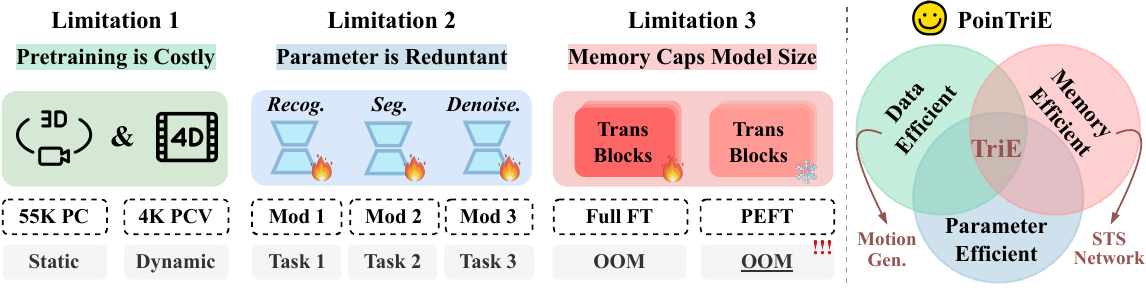}
  \caption{\textbf{Three key limitations existing in point cloud video learning.} \textcolor{green}{L1}: PC and PCV data is two orders of magnitude smaller than images. \textcolor{blue}{L2}: Using task-specific dedicated models leads to severe parameter redundancy. \textcolor{red}{L3}: Full fine-tuning often causes out-of-memory errors, and conventional PEFT methods still struggle to resolve this issue. To address these limitations, we propose PoinTriE, a unified framework that achieves strong performance along the three critical axes of efficiency.
  }
  \label{fig1}
\end{figure}

Despite these advancements, most of the prior art still relies on  additive structures, \ie, trainable adapter parameters within the pretrained model across various downstream tasks. However, as depicted in Fig.~\ref{fig1}, recent findings from NLP (Natural Language Processing) have revealed the redundancy of intermediate caches retained by each pretrained block during backpropagation in terms of GPU memory, underscoring the need to optimize the fine-tuning method to eliminate the substantial memory footprint~\cite{sung2022lst,fu2024dtl}. On the other hand, unlike images and language~\cite{radford2021learning}, collecting large-scale point cloud data requires enormous storage and computing resources~\cite{chang2015shapenet}, let alone dynamic point cloud videos (PCV)~\cite{HOI4d}. This implies that indiscriminately using a fixed training and fine-tuning paradigm may not fully tap into the potential of PFMs, motivating us to \textit{efficiently} adjust the existing paradigm to better accommodate each task.

In this paper, we naturally consider data-, parameter-, and memory-efficiency as the ideal optimization objectives, which we refer to as tri-efficient transfer learning (TriETL). As the focus pivots towards the TriETL problem, several new research questions emerge. First, since the underlying nature of PCV learning is still underexplored, the definition of TriETL remains unclear. For instance, defining it as a single-stage process or a two-stage process leads to a critical consideration, namely whether the fine-tuning network has the privilege to intervene in the effectiveness of customized PFMs. This leads us to our first key question: (1) How can we appropriately and formally define TriETL? {(§\ref{3.1})}. Second, once the definition of TriETL is established, a more essential question emerges. (2) How can we design an intuitive pipeline derived from this definition? (§\ref{data eff},~\ref{para method}).

To address these questions, we first theoretically analyze the TriETL problem in 3D vision. Our findings confirm the significance of TriETL and further reveal two key insights: rather than blindly expanding the scale of the dataset, it is better to carefully explore the intrinsic supervision within existing data and its variants. Moreover, the negative impact of intermediate cache occupancy on memory persists in the point cloud domain. Based on these insights, we propose PoinTriE, a novel two-stage PFM optimized for our objectives. Building on the definition, we utilize rigid body transformation to generate pseudo-motion trajectories, which comprise a series of variants derived from the same point cloud data. The Geometric-Motion Duality Network (GMD Net) first maps trajectory features and performs contrastive learning with their 2D projections and text corpora, while the subsequent dual-branch head assists in motion prior injection from the perspectives of regression and uncertainty. When it comes to fine-tuning, we freeze the backbone and adaptively learn the Spatio-temporal Side Network (STS Net) for each task in a ladder-side manner using a gradient flow masking strategy, thereby unlocking the full potential of PFM and low-rank linear transformation (LoRA)~\cite{hu2022lora} units. Our main contributions are fourfold.
\begin{itemize}
    \item To the best of our knowledge, we are the first to investigate the TriETL problem in 3D vision and provide the corresponding theoretical proofs.
    \item Building upon our insights into point cloud video learning, we propose a dedicated TriETL framework, termed PoinTriE, which can provide a baseline for further research from either a unified or an independent perspective.
    \item We present the GMD Network for pretraining, which introduces pseudo-motion trajectories to augment under-utilized static point cloud data for multi-modal contrastive learning. Then we allocate LoRA units to STS blocks via the gradient flow masking strategy, resulting in significantly fewer trainable parameters and lower GPU memory usage.
    \item Extensive experiments demonstrate that PoinTriE exhibits superior performance, \eg, it attains $94.37\%$ average accuracy on MSR-Action3D and $84.11\%$ mIoU on Synthia 4D, with gains of 1.05pp and 0.95pp over baselines.
\end{itemize}

\section{Related Work}
\subsection{Static\&Dynamic Point Cloud Pretraining}
Within the domain of point cloud pretraining, self-supervised strategies are primarily divided into contrastive and generative frameworks. The contrastive framework aims to maximize agreement between different augmentations of the same point cloud while minimizing similarity across distinct samples. Initially proposed for 2D images~\cite{radford2021learning,chen2020simple,shao2026unidef}, it has been effectively extended to 3D data~\cite{han2025rethinking,zhang2026pointcot,zhang2026cmhanet,wang2026pointrft} and 4D data~\cite{shen2023pointcmp,zhang2026diffusion,guo2026mantis,shao2026wavelet}. For instance, PointContrast~\cite{PointContrast2020} was an early milestone, followed by CPR~\cite{cpr}, PointCMP~\cite{shen2023pointcmp}, and PointCPSC~\cite{cpsc}, which improved feature extraction via various priors. Multimodal extensions further enrich this line of work.  CrossPoint~\cite{afham2022crosspoint}, CrossVideo~\cite{liu2024crossvideo} and VG4D~\cite{deng2024vg4d} leverage modality correspondences to boost discriminability, while RECON~\cite{qi2023contrast}, PTM~\cite{10604912ptm} and HyperPoint~\cite{SUN2026112800} embed contrastive objectives into generative pipelines to mitigate overfitting in Transformer-based methods. On the other hand, several advances focus on masked reconstruction. Inspired by MAE~\cite{he2022masked}, PointMAE~\cite{pang2022masked} and PointM2AE~\cite{m2ae} simplify training to capture geometric details. Subsequent works, such as JointMAE~\cite{guo2023joint}, PointGPT~\cite{chen2024pointgpt}, and PointDif~\cite{zheng2024pointdif}, extend the concepts of multi-modal integration, GPT, and Diffusion frameworks to pretraining, respectively. Despite the effectiveness of MaST-Pre~\cite{shen2023masked} and M2PSC~\cite{han2024maskedm2psc} in 4D representations, we notice that existing methods possess two main limitations: (1) They require cumbersome and time-consuming data collection, which is not feasible for most laboratories and companies. (2) All of them learn motion directly from the decoded features. As noted in PointRAE~\cite{pointraeliu2023regress} and PointSRA~\cite{wei2026point}, such pretext tasks inadvertently push the encoder to learn shortcuts, leading to poor transferability for motion-aware downstream 4D tasks. To tackle these limitations, we turn our attention to the field of static point clouds, where relatively mature metadata is already available~\cite{chang2015shapenet,uy2019revisiting}. Our encoder retains a strong capability in modeling spatial structures while emphasizing the capture of morphological priors to ensure the efficacy of transfer learning.

\subsection{Efficient Transfer Learning}
As highlighted above, PFMs are iterating at an unprecedented pace in terms of structure, method, and applications. This rapid evolution has fueled the PEFT community to become equally dynamic. Under the topic of efficient transfer learning (ETL), PEFT can be finely categorized into four schemes: selective, additive, prompt, and reparameterization. 3D-specific methods, such as IDPT~\cite{zha2023instanceidpt}, DAPT~\cite{zhou2024dynamicdapt}, PointPEFT~\cite{pointpefttang2024point}, and GAPrompt~\cite{ai2025gaprompt}, have achieved efficient adaptation to the unique requirements of 3D vision. PointCSA~\cite{csa} pioneered the bridging of 3D and 4D domains via additive adapters, making cross-modal tuning possible. PointATA~\cite{sun2026alignadaptrethinkingparameterefficient} decomposed fine-tuning into two stages, narrowing inter-domain discrepancy while mitigating overfitting risk. Nevertheless, parameter efficiency is not the community’s sole priority as data and memory efficiency are equally indispensable. LST~\cite{sung2022lst} and DTL~\cite{fu2024dtl} have set a precedent in the fields of language and 2D vision. They utilize side tuning to fuse intermediate information from the backbone network, thereby saving substantial memory. In this work, we hypothesize that the pretraining data scale and memory demands for special 4D scenarios may also play a crucial role, leading us to simultaneously incorporate data- and memory-efficiency into our focus. Furthermore, we extend TriETL to PFM architecture and demonstrate its practical usage on multiple high-dimensional vision tasks for the first time.

\section{Method}
\subsection{Analysis and Formulation of TriETL}
\label{3.1}
We first explore the underlying nature of TriETL and its formal definition to address question (1). Intuitively, TriETL can be defined in two distinct forms: a single-stage process (where data-, parameter-, and memory-efficiency are all achieved during pretraining) or a two-stage process (where these three objectives are allocated to the pretraining and fine-tuning processes, respectively). These two definitions differ in whether the fine-tuning network is capable of undertaking some of these optimization objectives. If we can prove that these three properties may be mutually exclusive under the single-stage condition, we can thereby prove by contradiction that multi-stage optimization is the appropriate definition.

Given the well-established scaling laws for upstream pretraining performance in prior work, we adopt a simple yet empirically validated formulation as the theoretical foundation for our analysis. Specifically, in visual transfer learning scenarios with limited labeled data, the downstream task error rate $E$ satisfies the following power-law scaling relation, which is adapted and generalized from Yang et al. (2025)~\cite{yang2025scaling}:
    \begin{equation}
    E\left(\mathcal{D}_p, \mathcal{M}, \mathcal{D}_f\right) = E_\infty + \frac{\mathcal{D}_p^{-\alpha}}{\lambda_p} + \frac{\mathcal{M}^{-\beta}}{\lambda_m} + \frac{\mathcal{D}_f^{-\gamma}}{\lambda_f},
    \label{eq:error_decomposition}
    \end{equation}
where all symbols are rigorously defined as follows. \( E_{\infty} \in \mathbb{R}_{+} \) is a constant representing the irreducible error of the downstream task and a larger \( E \) indicates worse performance in downstream transfer learning. $\mathcal{D}_p$ denotes the pretraining data size (\textit{corr.} \textcolor{green}{data efficiency}), $\mathcal{M}$ represents the model size (\textit{corr.} \textcolor{blue}{parameter} and \textcolor{red}{memory efficiency}), $\mathcal{D}_f$ denotes the fine-tuning data size, and $\lambda$ determines the characteristic scale of
each dimension.

To simplify the theoretical analysis, we define the constant term $\mathcal{C}_f = \frac{\mathcal{D}_f^{-\gamma}}{\lambda_f}$, which absorbs the fixed fine-tuning data contribution. We first introduce two axioms that underpin our subsequent theorem, formalizing the inherent trade-offs between model parameters and memory usage:

\noindent\textbf{Axiom 1} (Parameter-Memory Monotonicity). For any pretrained model, the memory usage $O(\mathcal{M})$ is a strictly increasing, continuously differentiable function of the model size $\mathcal{M}$, \ie, $O(\mathcal{M}) = j \cdot \mathcal{M} + o(\mathcal{M})$ for some constant $j$, where $o(\mathcal{M})$ is the lower-order term and negligible for large $\mathcal{M}$. Thus, a reduction in $\mathcal{M}$ implies a strict decrease in $O(\mathcal{M})$, and vice versa.

\noindent\textbf{Axiom 2} (Dual-Constraint Performance Degradation). Let $\mathcal{D}_p \leq \mathcal{D}_{p,\text{max}}$ and $\mathcal{M} \leq \mathcal{M}_{\text{max}}$ denote constraints on pretraining data size and model size, respectively. Then, the downstream error $E(\mathcal{D}_p, \mathcal{M}, \mathcal{C}_f)$ is strictly larger than the error achieved when at least one constraint is relaxed.

\noindent\textbf{Theorem} (Constrained Pretraining Feasibility). Let $E_\tau \in \mathbb{R}_{++}$ be the target downstream error threshold, satisfying $E_\tau \geq E_\infty$. If the pretraining data size is constrained to a finite upper bound $\mathcal{D}_p = \mathcal{D}_p^*$ (\ie, $\mathcal{D}_p \leq \mathcal{D}_p^* < +\infty$), then the model size $\mathcal{M}$ must be unconstrained to ensure that the feasible region of $\mathcal{M}$ is non-empty. 

\noindent\textbf{Proof}. We proceed via three key steps: error simplification, inequality derivation, and limit analysis.

\textit{Step 1: Error Function Simplification}. Define the constant term $\mathcal{C}_p^* = \frac{\mathcal{D}_p^{-\alpha,*}}{\lambda_p}$. Fix the pretraining data size and substitute $\mathcal{C}_p^*$ and $\mathcal{C}_f$ into Eq.~\eqref{eq:error_decomposition}, the downstream error simplifies to a univariate function of $\mathcal{M}$:

    \begin{equation}
    E(\mathcal{M}) = E_\infty + \mathcal{C}_p^* + \mathcal{C}_f + \frac{\mathcal{M}^{-\beta}}{\lambda_m},
    \label{eq:error_univariate}
    \end{equation}
where $E(\mathcal{M})$ is strictly decreasing in $\mathcal{M}$, since $\frac{\partial E(\mathcal{M})}{\partial \mathcal{M}} = -\frac{\beta \mathcal{M}^{-(\beta+1)}}{\lambda_m} < 0$.

\textit{Step 2: Inequality Derivation for Feasibility}. We require $E(\mathcal{M}) \leq E_\tau$. Substituting Eq.~\eqref{eq:error_univariate} into this inequality yields:

    \begin{equation}
    E_\infty + \mathcal{C}_p^* + \mathcal{C}_f + \frac{\mathcal{M}^{-\beta}}{\lambda_m} \leq E_\tau.
    \label{eq:error_inequality}
    \end{equation}

Combine the constant terms by defining $\Gamma = E_\tau - \left(E_\infty + \mathcal{C}_p^* + \mathcal{C}_f\right)$. By the definition of irreducible error, $E_\infty + \mathcal{C}_p^* + \mathcal{C}_f > E_\infty$, so $\Gamma$ can be either positive or negative. Isolating the term involving $\mathcal{M}$ in Eq.~\eqref{eq:error_inequality}, we obtain:

    \begin{equation}
    \mathcal{M}^{-\beta} \leq \lambda_m \cdot \Gamma.
    \label{eq:M_inequality}
    \end{equation}

\textit{Step 3: Limit Analysis and Feasibility}. Note that $\mathcal{M} \in \mathbb{R}_{++}$ and $\beta, \lambda_m \in \mathbb{R}_{++}$, so $\mathcal{M}^{-\beta} = \frac{1}{\mathcal{M}^\beta} > 0$ for all finite $\mathcal{M}$. We take the limit of $E(\mathcal{M})$ as $\mathcal{M} \to +\infty$. By the limit property of negative power functions, we obtain:

    \begin{equation}
    \lim_{\mathcal{M} \to +\infty} E(\mathcal{M}) = E_\infty + \mathcal{C}_p^* + \mathcal{C}_f + \lim_{\mathcal{M} \to +\infty} \frac{\mathcal{M}^{-\beta}}{\lambda_m} = E_\infty + \mathcal{C}_p^* + \mathcal{C}_f.
    \label{eq:error_limit}
    \end{equation}

Define $L = E_\infty + \mathcal{C}_p^* + \mathcal{C}_f$. By the Sign-Preserving Property of Limits~\cite{rudin1976principles}, for any $\zeta > 0$, there exists $\mathcal{M}^* \in \mathbb{R}_{++}$ such that for all $M \geq M^*$, $|E(M) - L| < \zeta$. Choosing $\zeta = \Gamma$, we obtain $E(\mathcal{M}) < L + \Gamma = E_\tau$ for all $\mathcal{M} \geq \mathcal{M}^*$. The feasible region of $\mathcal{M}$ is defined as $\mathbb{M} = \left\{ \mathcal{M} \in \mathbb{R}_{++} \mid E(\mathcal{M}) \leq E_\tau \right\}$. From the above analysis, $\{ \mathcal{M} \in \mathbb{R}_{++} \mid \mathcal{M} \geq \mathcal{M}^* \} \subseteq \mathbb{M}$, which implies $\mathbb{M} \neq \emptyset$. 

\noindent\textbf{Remark}. Theorem highlights a fundamental trade-off between pretraining data efficiency and parameter efficiency in limited transfer learning, which indicates that the above attributes cannot be simultaneously optimized in constrained pretraining. Motivated by this trade-off, we formally define the TriETL problem as a two-stage optimization framework.

\subsection{Data-Efficient 3D Pretraining}
\label{data eff}
\begin{figure}[tb]
  \centering
  \includegraphics[width=\linewidth]{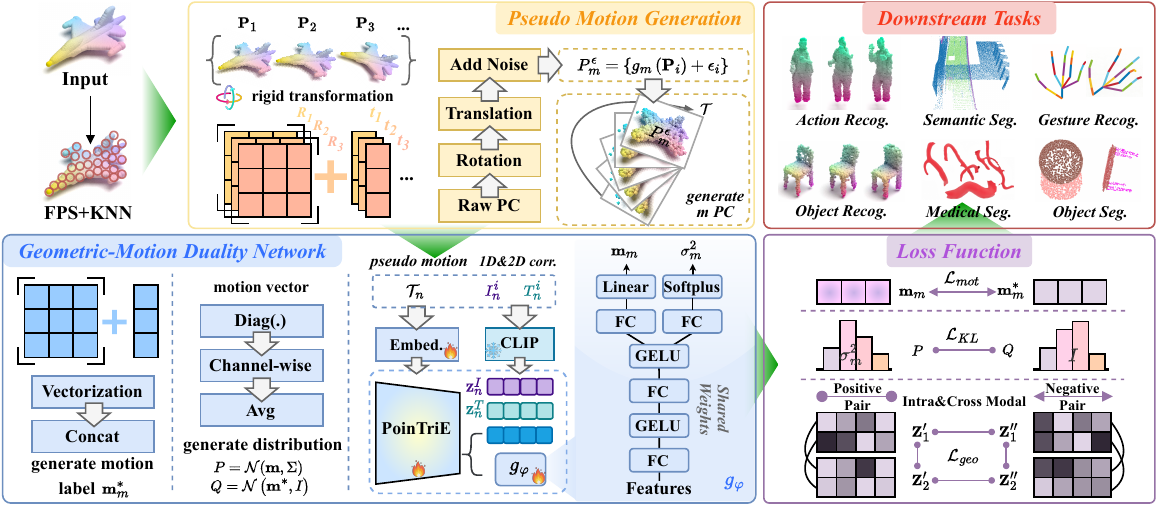}
  \caption{\textbf{The overall pretraining framework of PoinTriE.} The input first undergoes rigid transformations to augment the data scale. Motion prediction and KL divergence enhance sensitivity to fine-grained dynamic changes. Cross-modal alignment subsequently injects additional semantic priors to support various downstream tasks.
  }
  \label{fig2}
\end{figure}

\textbf{Pseudo Motion Generation.} To ensure that pseudo-motion adheres to the physical motion patterns of real PCVs, we predefine a set of rigid transformations. Given an input point cloud \( P \), we first construct local point cloud patches via Farthest Point Sampling (FPS) and K-Nearest Neighbors (KNN) clustering.  For the \( m \)-th rigid transformation \( g_m^r \in SE(3) \), we define the transformation as:
\begin{equation}
g_m^r\left(\mathbf{p}_i\right)=R_m \mathbf{p}_i+t_m,
\end{equation}
where \( R_m \in \mathbb{R}^{3 \times 3} \) is a rotation matrix with rotation angles constrained to \( \theta_m \in [-30^\circ, 30^\circ] \). \( t_m \in \mathbb{R}^3 \) is a translation vector whose components follow the uniform distribution \( t_{m,x}, t_{m,y}, t_{m,z} \sim \mathcal{U}(-0.2, 0.2) \) \underline{\textit{(cf Supp E)}}. 

For a single 3D point cloud \( P \), we apply \( M \) independent rigid transformations and add per-point Gaussian noise \( \epsilon_i \sim \mathcal{N}(0, \sigma^2 I) \), resulting in the noisy pseudo-motion point clouds \( P_m^\epsilon = \left\{ g_m^r(\mathbf{p}_i) + \epsilon_i \mid \mathbf{p}_i \in P \right\} \). By aggregating all pseudo-motion samples derived from the same input point cloud, we construct the final pseudo-motion trajectory set \( \mathcal{T} = \left\{ P, P_1^\epsilon, P_2^\epsilon, \ldots, P_M^\epsilon \right\} \).

\textbf{Geometric-Motion Duality Network (GMD Net)}.
For the pseudo-motion trajectory $\mathcal{T}=\left\{P, P_1^\epsilon, P_2^\epsilon, \ldots, P_M^\epsilon\right\}$ generated from a single-frame point cloud $P$, we divide the pseudo-motion samples $\left\{P_1^\epsilon, \ldots, P_M^\epsilon\right\}$ into the positive sample set as:
\begin{equation}
\mathcal{S}_{\mathrm{pos}}=\left\{P_1^\epsilon, P_2^\epsilon, \ldots, P_M^\epsilon\right\}.
\end{equation}

We take other point cloud and their trajectories within the batch as the negative sample set. We use our 3D point cloud encoder $f_\theta(\cdot)$ to encode point cloud features, obtaining sample-wise global features. For cross-modal alignment, we encode the image input $I_i$ and text corpora ${T}_i$ separately as follows:

\begin{equation}
\mathbf{z}_n^P=f_\theta\left(P_n\right), \quad \mathbf{z}_n^I=f_I\left(I_n\right), \quad \mathbf{z}_n^T=\operatorname{Avg}\left(f_T\left({T}_n\right)\right),
\end{equation}
where $f_I(\cdot)$ is the image encoder, $f_T(\cdot)$ is the text encoder, and $\operatorname{Avg}(\cdot)$ denotes the global average pooling of text features \underline{\textit{(cf Supp A.1)}}. 

Meanwhile, we vectorize the real rigid transformation parameters corresponding to the $m$-th pseudo motion to construct the real motion label,

\begin{equation}
\mathbf{m}_m^*=\left[\operatorname{vec}\left(R_m\right) ; t_m\right] \in \mathbb{R}^{12},
\end{equation}
where $\operatorname{vec}(\cdot)$ is the matrix column vectorization operation. This label has a one-to-one correspondence with the pseudo-motion point cloud $P_m^\epsilon$. 

In this manner, we introduce two Gaussian distributions to respectively model the true motion distribution and the predicted motion distribution. The true motion distribution is \( \mathcal{Q} = \mathcal{N}(\mathbf{m}_m^*, I) \), where the covariance matrix is the identity matrix \( I \), indicating that the true motion value is definite and unambiguous. The predicted motion distribution is \( \mathcal{P} = \mathcal{N}(\mathbf{m}_m, \Sigma_m) \), where \( \mathbf{m}_m \) denotes the predicted motion vector and \( \Sigma_m \) is the predicted diagonal covariance matrix.

\textbf{Loss Function.} 
The total loss of GMD Net is composed of the geometric invariance loss $\mathcal{L}_{\mathrm{geo}}$ and the motion consistency loss $\mathcal{L}_{\mathrm{mot}}$. The geometric invariance loss comprises the intra-modal point cloud contrastive loss and the cross-modal alignment loss. First, for different pseudo-motion samples of the same input point cloud, we encourage the encoder to output similar global representations. Given a positive sample pair \( \mathbf{z}_n^{t_1} \) and \( \mathbf{z}_n^{t_2} \), the contrastive loss is defined as:
\begin{equation}
\mathrm{l}\left(n, t_1, t_2\right)=-\log \frac{\exp \left(s\left(\mathbf{z}_n^{t_1}, \mathbf{z}_n^{t_2}\right) / \tau\right)}{\sum_{\substack{k=1 \\ k \neq n}}^B \exp \left(s\left(\mathbf{z}_n^{t_1}, \mathbf{z}_k^{t_1}\right) / \tau\right)+\sum_{k=1}^B \exp \left(s\left(\mathbf{z}_n^{t_1}, \mathbf{z}_k^{t_2}\right) / \tau\right)},
\end{equation}
where $B$ is the batch size, $\tau$ is the temperature coefficient, and $s(\cdot, \cdot)$ denotes the cosine similarity function. Furthermore, the intra-modal geometric invariance loss is given by:
\begin{equation}
\mathcal{L}_{\mathrm{intra}}=\frac{1}{2B} \sum_{n=1}^B\left[\mathrm{l}\left(n, t_1, t_2\right)+\mathrm{l}\left(n, t_2, t_1\right)\right].
\end{equation}

The cross-modal loss aligns 3D point cloud, image, and text features into a unified semantic space. It leverages the vision-language prior of the pretrained CLIP~\cite{radford2021learning} model to enhance the semantic consistency of geometric representations. We compute the weighted sum of all modal pairs with InfoNCE~\cite{Park2020FeatureLevelEKinfonce} loss:
\begin{equation}
\mathcal{L}_{\text {geo}}=\mathcal{L}_{\text {intra}}+\mathcal{L}_{(I,P)}+\mathcal{L}_{(S, P)}=\mathcal{L}_{\text {intra}}+\mathcal{L}_{\text {cross}}.
\end{equation}

On the other hand, we employ a lightweight dual mapping network \( g_\varphi \) to simultaneously predict the motion vector and its associated uncertainty. This network concurrently outputs the predicted motion vector and the diagonal elements of the predicted covariance. 

We use the L2 regression loss to enforce numerical consistency between the predicted motion vector \( \mathbf{m}_m \) and the ground-truth motion label \( \mathbf{m}_m^* \). To enhance the model's robustness against noise and point cloud perturbations, we constrain the predicted motion distribution to approximate the true motion distribution via KL divergence. The overall motion consistency loss is thus defined as:
\begin{equation}
\mathcal{L}_{\mathrm{mot}} = \frac{1}{M} \sum_{m=1}^M \left( \eta_1 \left\| \mathbf{m}_m - \mathbf{m}_m^* \right\|_2^2 + \eta_2 \mathbb{D}_{\mathrm{KL}} \left( \mathcal{N}\left( \mathbf{m}_m, \Sigma_m \right) \| \mathcal{N}\left( \mathbf{m}_m^*, I \right) \right) \right),
\end{equation}
where $\eta_1=1$ and $\eta_2=0.1$, $\Sigma_m=\operatorname{diag}\left(\sigma_{m, 1}^2, \ldots, \sigma_{m, 12}^2\right)$ is the predicted diagonal covariance matrix \underline{(\textit{cf Supp B.2})}.

These two types of information collectively form the supervision source for geometric-motion duality learning. The total loss of GMD Net is formulated as:
\begin{equation}
\mathcal{L}_{\mathrm{GMDP}}=\mathcal{L}_{\mathrm{geo}}+\delta \mathcal{L}_{\mathrm{mot}},
\end{equation}
where $\delta$ is a coefficient that balances the supervision of geometry and motion.

\subsection{Parameter- and Memory-Efficient 4D Transfer Learning}
\label{para method}
Despite the significant reduction in trainable parameters achieved by PEFT methods, the decrease in GPU memory usage is not proportional, rendering the fine-tuning of large-scale PFMs still challenging. To simplify the analysis, we assume the input PCV sample is \( \mathbf{h}^{(0)} \in \mathbb{R}^{F \times N \times 3} \), where $F$ is the length of the video and $N$ is the number of points per frame. We exclude bias terms in the hidden layers. For the \( l \)-th layer, the linear transformation, activation value and the output can be expressed as:
\begin{equation}
\mathbf{v}^{(l)}=\mathbf{W}^{(l)} \mathbf{h}^{(l-1)},\quad \mathbf{h}^{(l)}=\phi_l\left(\mathbf{v}^{(l)}\right), \quad \mathbf{o}=\mathbf{h}^{(l)},
\end{equation}
where \( \mathbf{W}^{(l)} \in \mathbb{R}^{h_l \times h_{l-1}} \) denotes the weight matrix. \( \phi_l\) is an element-wise differentiable activation function.

Formally, we define the objective function as \( \mathcal{J} = \mathcal{L}(\mathbf{o}, y) + \frac{\mu}{2} \sum_{k=1}^l \left\| \mathbf{W}^{(k)} \right\|_F^2 \), where \( \mathcal{L}(\cdot, \cdot) \) is the differentiable task loss, \( y \) denotes the label, \( \mu > 0 \) is the regularization coefficient, and \( \|\cdot\|_F \) represents the Frobenius norm. 

The gradient of the objective function with respect to the output layer can be uniformly expressed as:
\begin{equation}
\boldsymbol{\delta}^{(l)}=\frac{\partial \mathcal{J}}{\partial \mathbf{v}^{(l)}}=\frac{\partial \mathcal{L}}{\partial \mathbf{o}} \odot \phi_l^{\prime}\left(\mathbf{v}^{(l)}\right), \quad \frac{\partial \mathcal{J}}{\partial \mathbf{W}^{(l)}}=\boldsymbol{\delta}^{(l)}\left(\mathbf{h}^{(l-1)}\right)^{\top}+\mu \mathbf{W}^{(l)},
\end{equation}
where \( \odot \) denotes the Hadamard product.

Notably, \( \phi_l'(\mathbf{v}^{(l)}) \) and the outer product \( \boldsymbol{\delta}^{(l)}(\mathbf{h}^{(l-1)})^\top \) depend on the forward propagation intermediate variables \( \mathbf{v}^{(l)} \) and \( \mathbf{h}^{(l-1)} \), respectively. By induction, gradient computation for each layer \( k \in \{1, \ldots, l\} \) relies on at least one forward intermediate variable, either \( \mathbf{v}^{(k)} \) or \( \mathbf{h}^{(k-1)} \). Thus, even when only a subset of parameters is updated, PEFT methods incur nearly  identical memory cost as full fine-tuning due to the need to cache these intermediate variables.

\begin{figure}[tb]
  \centering
  \includegraphics[width=\linewidth]{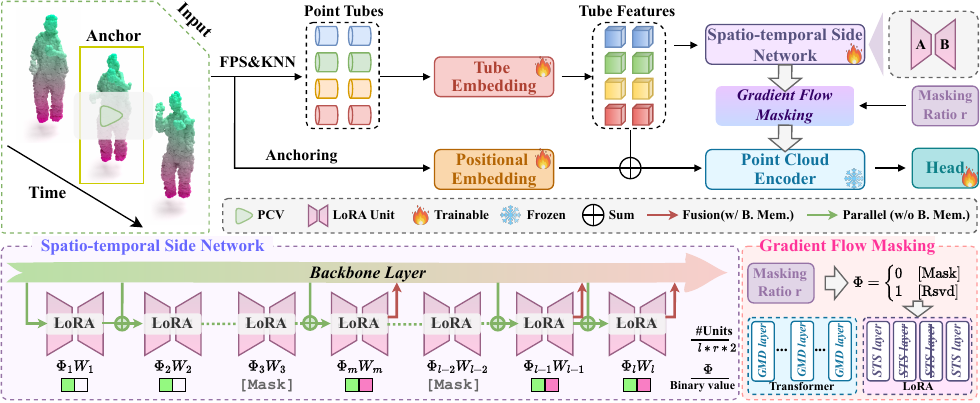}
  \caption{\textbf{The overall PCV fine-tuning framework in PoinTriE.} First, anchor centers are sampled and embedded via FPS and KNN. Then, we freeze the backbone and insert the STS Net in parallel. At each block, the STS Net applies LoRA units to the input features to extract task-specific side information \textcolor{green1}{(green arrows)}. Starting from block $m$, this side information is fused back to the outputs of the block \textcolor{red1}{(red arrows)} to adapt the backbone representations for downstream tasks. During fine-tuning, we adopt a gradient flow masking strategy to further reduce the parameters of the STS Net, which is controlled by masking ratio.
  }
  \label{fig3}
\end{figure}

To address this memory bottleneck, we propose our Spatio-Temporal Side Network. As illustrated in Fig.~\ref{fig3}, the core idea of STS Net is to decouple the weight update of lightweight auxiliary modules from the backbone network. This design drastically reduces the intermediate variables $\mathbf{v}^{(l)}$ and $\mathbf{h}^{(l-1)}$ that must be cached for backpropagation. As a result, STS Net is not only parameter-efficient but also substantially reduces the GPU memory footprint for fine-tuning PFMs.

Specifically, given a 3D backbone with $C$ blocks, the forward computation can be written as:
\begin{equation}
\mathbf{o}=b_C\left(b_{C-1}\left(\ldots b_1\left(\mathbf{h}^{(\mathbf{0})}\right)\right)\right),
\end{equation}
where $b_i$ denotes the $i$-th backbone block \underline{\textit{(cf Supp A.2/D.2)}}. 

STS Net comprises $C$ low-rank linear transformation~\cite{hu2022lora} matrices, with one matrix inserted into each block to extract task-specific information. Let $w_i=a_i c_i \in \mathbb{R}^{d \times d}$ be the low-rank weight matrix for the $i$-th block, where $a_i \in \mathbb{R}^{d \times d^{\prime}}, c_i \in \mathbb{R}^{d^{\prime} \times d}$, and $d^{\prime} \ll d$. We initialize $a_i$ from a uniform distribution and set $c_i$ to zero. Empirically, we find that fully-equipped LoRA units suffer from high redundancy and tend to overfit on downstream datasets. To alleviate this issue, we further introduce a gradient flow masking strategy to reduce redundancy. This strategy randomly removes a subset of LoRA units, controlled by a predefined masking ratio, which improves model generalization and cross-modal transfer ability. Starting from the $D$-th block, the aggregated task-specific information $\mathbf{h}^{(i+1)}$ is used to adapt the backbone features $\mathbf{v}^{(i+1)}$ for downstream tasks via a residual adaptation. 
\begin{equation}
\mathbf{h}_{i+1}^{\prime}=\left\{\begin{array}{ll}
\Phi(\mathbf{h}_{i+1}+\phi\left(\mathbf{v}_{i+1}\right)) & \text { if } i \geq D \\
\Phi \mathbf{h}_{i+1} & \text { otherwise }
\end{array}, \quad \Phi= \begin{cases}0 & {[\text { Mask }]} \\
1 & {[\text { Rsvd }]}\end{cases}.\right.
\end{equation}

\section{Experiments}
Extensive experiments are conducted on three point cloud video benchmarks: MSR-Action3D~\cite{5543273}, SHREC'17~\cite{desmedt:hal-01563505}, and Synthia 4D~\cite{choy20194d}. Our backbone is first pretrained and then fine-tuned under the proposed TriETL paradigm. We evaluate the performance of our method on three downstream tasks including action recognition, gesture recognition, and semantic segmentation. We further carry out comprehensive ablation studies to verify the effectiveness of each component within our method \underline{\textit{(cf Supp B-D for more experiments and ablation studies)}}.

\begin{table}[t]
\caption{\textbf{Action recognition accuracy (\%) on MSR-Action3D.} SL: Supervised Learning; PD: Pretraining Data; Para: Tunable Parameters; Mem: Memory; FFT: Full Fine-tuning; MSR: MSR-Action3D; SN: ShapeNet; F: Frames; Avg: Average Performance Our method is highlighted in \colorbox{myhl}{light blue}. Same for subsequent tables.
}
\label{tab1}
\centering
\definecolor{mainbg}{HTML}{E1F4FC}
\definecolor{sidebg}{HTML}{fdc7cf}
\resizebox{\linewidth}{!}{
\begin{tabular}{lcccccccccc}
\hline
\textbf{Methods}                                                                    & \textbf{Ref} & \textbf{Paradigm} & \textbf{PD} & \textbf{Para} & \textbf{Mem} & \textbf{8 F.} & \textbf{12 F.} & \textbf{16 F.} & \textbf{24 F.} & \textbf{Avg} \\ \hline
MeteorNet~\cite{liu2019meteornet}                           & ICCV    & SL                & N/A         & N/A           & N/A          & 81.14         & 86.53          & 88.21          & 88.50          & 86.09        \\
PSTNet~\cite{fan2021pstnet}                                   & ICLR     & SL                & N/A         & N/A           & N/A          & 83.50         & 87.88          & 89.90          & 91.20          & 88.12        \\
P4Trans~\cite{fan21p4transformer}                       & CVPR     & SL                & N/A         & N/A           & N/A          & 83.17         & 87.54          & 89.56          & 90.94          & 87.80        \\
Kinet~\cite{zhong2022no}                                      & CVPR    & SL                & N/A         & N/A           & N/A          & 83.84         & 88.53          & 91.92          & 93.27          & 89.39        \\
PPTr~\cite{wen2022point}                                      & ECCV     & SL                & N/A         & N/A           & N/A          & 84.02         & 89.89          & 90.31          & 92.33          & 89.13        \\
LeaF~\cite{10377208}                                          & ICCV    & SL                & N/A         & N/A           & N/A          & 84.50         & N/A            & 91.50          & 93.84          & N/A          \\
PST-Trans~\cite{9740525}                                & TPAMI   & SL                & N/A         & N/A           & N/A          & 83.97         & 88.15          & 91.98          & 93.73          & 89.45        \\
X4D-Scene~\cite{jing2024x4d}                          & AAAI   & SL                & N/A         & N/A           & N/A          & 86.47         & N/A            & 92.56          & 93.90          & N/A          \\
MAMBA4D~\cite{liu2025mamba4d}                                 & CVPR     & SL                & N/A         & N/A           & N/A          & N/A           & N/A            & N/A            & 92.68          & N/A          \\ \hline
CPR~\cite{cpr}                                                & AAAI    & FFT               & MSR.        & 100\%         & N/A          & 86.53         & 91.00          & 92.15          & 93.03          & 90.67        \\
C2P~\cite{zhang2023complete}                                  & CVPR    & FFT               & MSR.        & 100\%         & N/A          & 87.16         & N/A            & 91.89          & 94.76          & N/A          \\
PointCMP~\cite{shen2023pointcmp}                              & CVPR   & FFT               & MSR.        & 100\%         & N/A          & 89.56         & 91.58          & 92.26          & 93.27          & 91.66        \\
PointCPSC~\cite{cpsc}                                         & ICCV   & FFT               & MSR.        & 100\%         & N/A          & 88.89         & 90.24          & 92.26          & 92.68          & 91.01        \\
MaST-Pre~\cite{shen2023masked}                                & ICCV    & FFT               & MSR.        & 100\%         & N/A          & N/A           & N/A            & N/A            & 94.08          & N/A          \\ \hline
PointCSA~\cite{csa}                                           & CVPR    & PEFT              &SN.    & 3.4\%         & 26.4         & 90.72         & 92.39          & 93.38          & 94.77          & 92.81        \\
PointATA\cite{sun2026alignadaptrethinkingparameterefficient} & TMM    & PEFT              &SN.   & 2.8\%         & 28.5         & 91.28         & 92.68          & 94.07          & 95.28          & 93.32        \\
\rowcolor{mainbg} PoinTriE                                         & -            & TriETL            &SN.    & 2.2\%         & 9.8          & 91.28         & 93.37          & 95.62          & 97.21          & 94.37        \\ \hline
\end{tabular}}
\end{table}

\subsection{Action Recognition}
\textbf{Dataset.} We conduct experiments on MSR-Action3D~\cite{5543273}, which contains 567 depth videos covering 20 action classes. Each video contains approximately 40 frames on average. Following previous work~\cite{sun2026alignadaptrethinkingparameterefficient,csa}, we split the dataset into 270 training videos and 297 test videos. During fine-tuning, only point coordinates are utilized, and each frame is randomly sampled to 2048 points. Videos are partitioned into clips of 8, 12, 16, or 24 frames, and the final video-level accuracy is obtained by averaging clip-level predictions.

\textbf{Comparison Results.} To ensure a fair comparison, we primarily compare our method with those adopting similar training paradigms, specifically PointCSA~\cite{csa} and PointATA~\cite{sun2026alignadaptrethinkingparameterefficient}. As illustrated in Tab.~\ref{tab1}, our method achieves the highest action recognition accuracy, surpassing both supervised and fully fine-tuned counterparts. This demonstrates that the proposed pretext tasks, which emphasize geometric invariance and motion consistency, enable the model to effectively capture spatio-temporal information in point cloud videos. Such a design also provides substantial benefits when applying static models to dynamic downstream tasks. Notably, PoinTriE maintains its superiority over existing methods under the 12, 16, and 24-frame settings, yielding an average improvement of 1.05\%, which attests to the robustness of our method across diverse temporal resolutions.

\subsection{Gesture Recognition}
\textbf{Dataset.} We utilize SHREC'17~\cite{desmedt:hal-01563505} for gesture recognition. The dataset contains 2800 videos covering 28 gesture categories. Following~\cite{shen2023masked}, we split the dataset into 1960 training videos and 840 test videos. 

\textbf{Comparison Results.} As illustrated in Tab.~\ref{tab2}, our method outperforms FFT and SL methods by a substantial margin, achieving the accuracy of 96.5\%. Since gesture recognition inherently demands fine-grained modeling of geometric semantics, existing PEFT methods heavily rely on powerful pre-trained backbones such as PointBERT~\cite{yu2021pointbert}, PointMAE~\cite{pang2022masked} and PointGPT~\cite{chen2024pointgpt}, which lack dedicated designs for joint feature learning. This limitation is precisely addressed by our GMD Net. By incorporating the STS Net in parallel, PoinTriE further boosts the performance of the backbone model. With fine-tuning on limited target-domain data, PoinTriE sheds new light on point cloud video learning.

\begin{table}[t]
\centering
\begin{minipage}{0.5\textwidth}
\centering
\caption{\textbf{Gesture recognition accuracy (\%) on SHREC'17.}}
\label{tab2}
\captionsetup{width=\textwidth} 
\resizebox{\linewidth}{!}{
\begin{tabular}{lccc}
\toprule
Methods               & Reference & Paradigm & Acc. \\ \hline
PLSTM-b~\cite{9157795}  & CVPR2020  & SL       & 87.6     \\
PLSTM-e~\cite{9157795}  & CVPR2020  & SL       & 93.5     \\
PLSTM-P~\cite{9157795}   & CVPR2020  & SL       & 93.1     \\
PLSTM-m~\cite{9157795} & CVPR2020  & SL       & 94.7     \\
PLSTM-l~\cite{9157795}  & CVPR2020  & SL       & 93.5     \\
Kinet~\cite{zhong2022no}       & CVPR2022  & SL       & 95.2     \\ \hline
PointCMP~\cite{shen2023pointcmp}      & CVPR2023  & FFT      & 93.3     \\
MaST-Pre~\cite{shen2023masked}    & ICCV2023  & FFT      & 92.4     \\
M2PSC~\cite{han2024maskedm2psc} & ECCV2024 &FFT &92.8 \\
\hline
PointCSA~\cite{csa}        & CVPR2025         & PEFT     & 95.2     \\
PointATA~\cite{sun2026alignadaptrethinkingparameterefficient}         & TMM2026         & PEFT     & 95.5     \\
\rowcolor{mainbg}
PoinTriE        & -         & TriETL   & 96.5     \\ \bottomrule
\end{tabular}}
\end{minipage}
\hfill 
\begin{minipage}{0.48\textwidth}
\centering
\caption{\textbf{4D semantic segmentation results (\%) on Synthia 4D.}}
\label{tab3}
\captionsetup{width=\textwidth} 
\resizebox{\linewidth}{!}{
\begin{tabular}{lccc}
\toprule
Methods                  & Reference    & Frame & mIoU\\ \hline
3D MinkNet14~\cite{choy20194d}   & CVPR2019 & 1     & 76.24     \\
4D MinkNet14~\cite{choy20194d}    & CVPR2019  & 3     & 77.46     \\ \hline
MeteorNet-M~\cite{liu2019meteornet}    & ICCV2019  & 2     & 81.47     \\
MeteorNet-L~\cite{liu2019meteornet}   & ICCV2019  & 3     & 81.80     \\
PSTNet~\cite{fan2021pstnet}         & ICLR2021 & 3     & 82.24     \\
P4Trans~\cite{fan21p4transformer} & CVPR2021  & 1     & 82.41     \\
P4Trans~\cite{fan21p4transformer}   & CVPR2021  & 3     & 83.16     \\
PST-Trans~\cite{9740525} & PAMI2022 & 1     & 82.92     \\
PST-Trans~\cite{9740525} & PAMI2022  & 3     & 83.95     \\
MAMBA4D~\cite{liu2025mamba4d}     & CVPR2025  & 3     & 83.35     \\ 
PointATA~\cite{sun2026alignadaptrethinkingparameterefficient}                & TMM2026         & 1     & 82.92     \\
PointATA~\cite{sun2026alignadaptrethinkingparameterefficient}                & TMM2026        & 3     & 84.06    \\
\rowcolor{mainbg}
PoinTriE                & -         & 3     & 84.11    \\ 
\bottomrule
\end{tabular}}
\end{minipage}
\end{table}

\subsection{Semantic Segmentation}
\textbf{Dataset.} To verify that PoinTriE handles point-level tasks, we apply it to 4D semantic segmentation on Synthia 4D~\cite{choy20194d}. This dataset includes six driving sequences where both objects and cameras are in motion. Each frame provides four stereo RGB-D images captured from a vehicle-mounted platform. Following~\cite{liu2025mamba4d,sun2026alignadaptrethinkingparameterefficient}, we reconstruct point cloud videos and use the standard train–validation–test split with 19888 training frames, 815 validation frames, and 1886 test frames. Consistent with~\cite{liu2019meteornet}, we use clips formed by three consecutive frames. While single-frame segmentation is feasible, modeling temporal correlations effectively enhances scene understanding, segmentation accuracy, and robustness to noise. 

\textbf{Comparison Results.} We report performance using mean Intersection over Union (mIoU) in Tab.~\ref{tab3}. Without adopting task-specific modules, PoinTriE consistently outperforms previous CNN, Transformer and Mamba models, achieving a 0.95 mIoU improvement over P4Transformer~\cite{fan21p4transformer}. While our performance is comparable to existing PEFT methods, our framework offers superior practicality by drastically reducing GPU memory usage, which is essential for lab-scale experimental environments.

\textbf{Visualization Samples.} We present two segmentation results in Fig.~\ref{fig6}. Our method demonstrates strong discriminative capability in scenarios with high semantic ambiguity, such as dense clusters of multiple objects.

\begin{table}[t]
\centering
\begin{minipage}{0.48\textwidth}
\centering
\caption{\textbf{Ablation studies of the pretraining objectives \underline{w/o fine-tuning}.}}
\label{tab4-1} 
\captionsetup{width=\textwidth}  
\begin{tabular}{lccccc}
\toprule
\multirow{2}{*}{Model} & \multirow{2}{*}{Motion} & \multicolumn{2}{c}{Contrastive}                 & \multirow{2}{*}{KL}    & \multirow{2}{*}{MSR.} \\ \cline{3-4}
                       &                         & Intra                  & Cross                  &                        &                       \\ \hline
A0                     & \ding{55}  & $\checkmark$ & \ding{55} & \ding{55} & 83.50                 \\
A1                     & $\checkmark$            & $\checkmark$           & \ding{55} & \ding{55} & 93.03                 \\
A2                     & $\checkmark$  & $\checkmark$ & $\checkmark$           & \ding{55} & 94.42                 \\
\rowcolor{mainbg}
A3                     & $\checkmark$            & $\checkmark$           & $\checkmark$           & $\checkmark$           & 95.42  \\ \bottomrule
\end{tabular}
\end{minipage}
\hfill 
\begin{minipage}{0.48\textwidth}
\centering
\caption{\textbf{Ablation studies of the fine-tuning components.}}
\label{tab4-2} 
\captionsetup{width=\textwidth}
\begin{tabular}{lccccc}
\toprule
\multirow{2}{*}{Model} & \multirow{2}{*}{Adapter} & \multicolumn{2}{c}{Schemes}                     & \multirow{2}{*}{Mask} & \multirow{2}{*}{MSR.} \\ \cline{3-4}
                       &                          & Side                   & Additive               &                             &                       \\ \hline
B0                     & \ding{55}   & \ding{55} & \ding{55} & \ding{55}      & 95.42                 \\
B1                     & $\checkmark$             & \ding{55}         & $\checkmark$ & \ding{55}      & 96.16                 \\
B2                     & $\checkmark$   & $\checkmark$ & \ding{55}  & \ding{55}      & 96.86                 \\
\rowcolor{mainbg}
B3                     & $\checkmark$             & $\checkmark$           & $\checkmark$           & $\checkmark$                & 97.21  \\ \bottomrule
\end{tabular}
\end{minipage}
\end{table}

\begin{figure}[tb]
\centering
\begin{minipage}{0.47\textwidth}
\centering
\includegraphics[width=\linewidth]{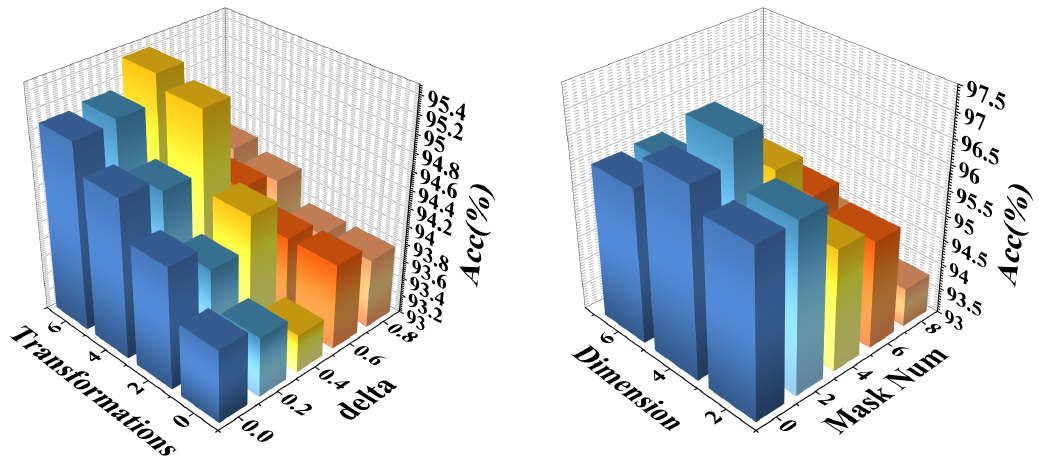}  
\caption{\textbf{Ablation studies of the pretraining\textit{ (left) }and fine-tuning \textit{(right) }hyperparameters.}}
\label{fig4}  
\end{minipage}
\hfill  
\begin{minipage}{0.5\textwidth}
\centering
\includegraphics[width=\linewidth]{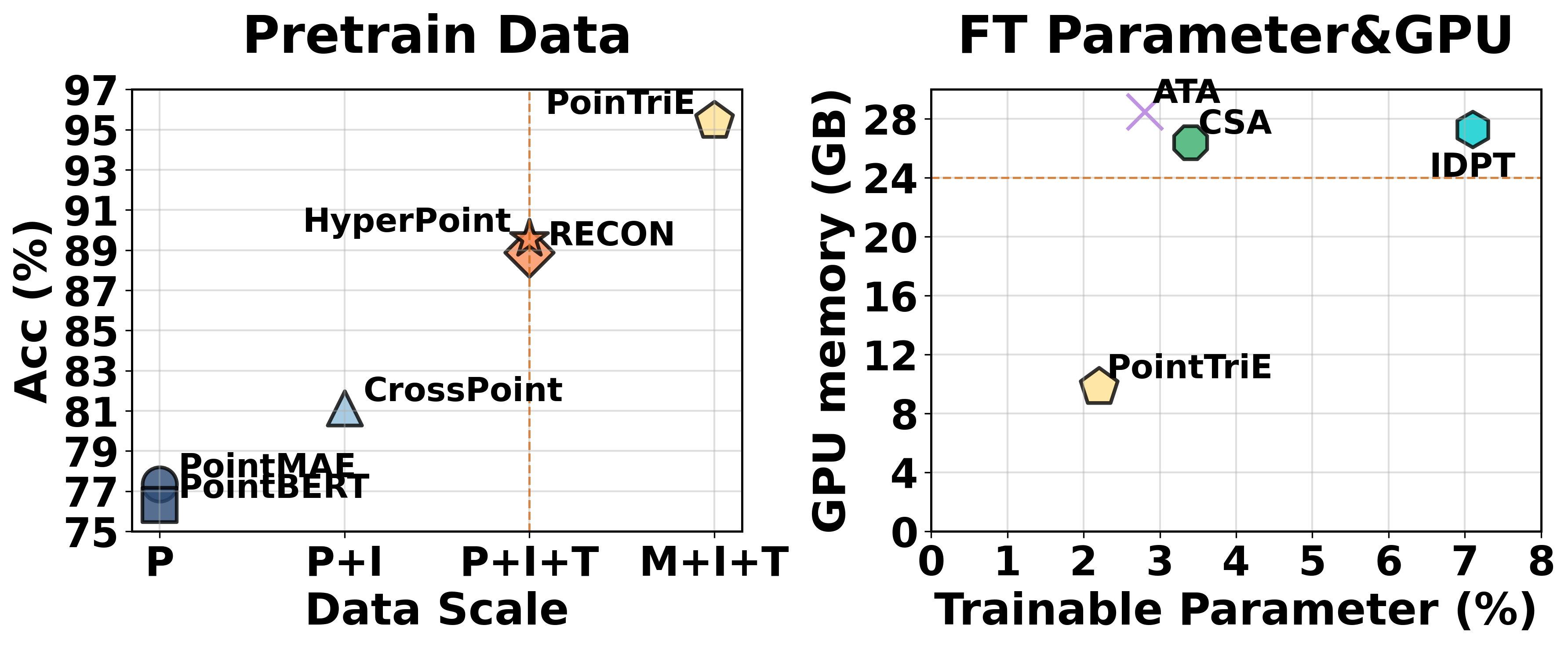}  
\caption{\textbf{Ablation studies of pretrained data \textit{(left)}, GPU memory, and tunable parameter ratio \textit{(right)}.}}
\label{fig5}  
\end{minipage}
\end{figure}

\subsection{Ablation Studies}
\label{sec:line-numbering}
\textbf{Pretraining Objectives.} Tab.~\ref{tab4-1} reports the impact of each pretraining objective on downstream tasks. The results indicate that static point cloud contrastive learning alone~\cite{PointContrast2020} fails to capture effective action patterns, leading to performance below that of the 4D-specific backbone. When the motion consistency objective is introduced, action recognition performance is substantially improved, which aligns with our core motivation. Incorporating KL divergence and more discriminative supervision, our method finally outperforms the 4D counterparts. 

\textbf{Fine-Tuning Components.} After validating the effectiveness of our pretrained backbone, we further analyze the contribution of each component in the fine-tuning pipeline. As shown in Tab.~\ref{tab4-2}, PointCSA~\cite{csa} (B1), a representative 

\begin{wrapfigure}{r}{0.5\textwidth}
\centering
\includegraphics[width=\linewidth]{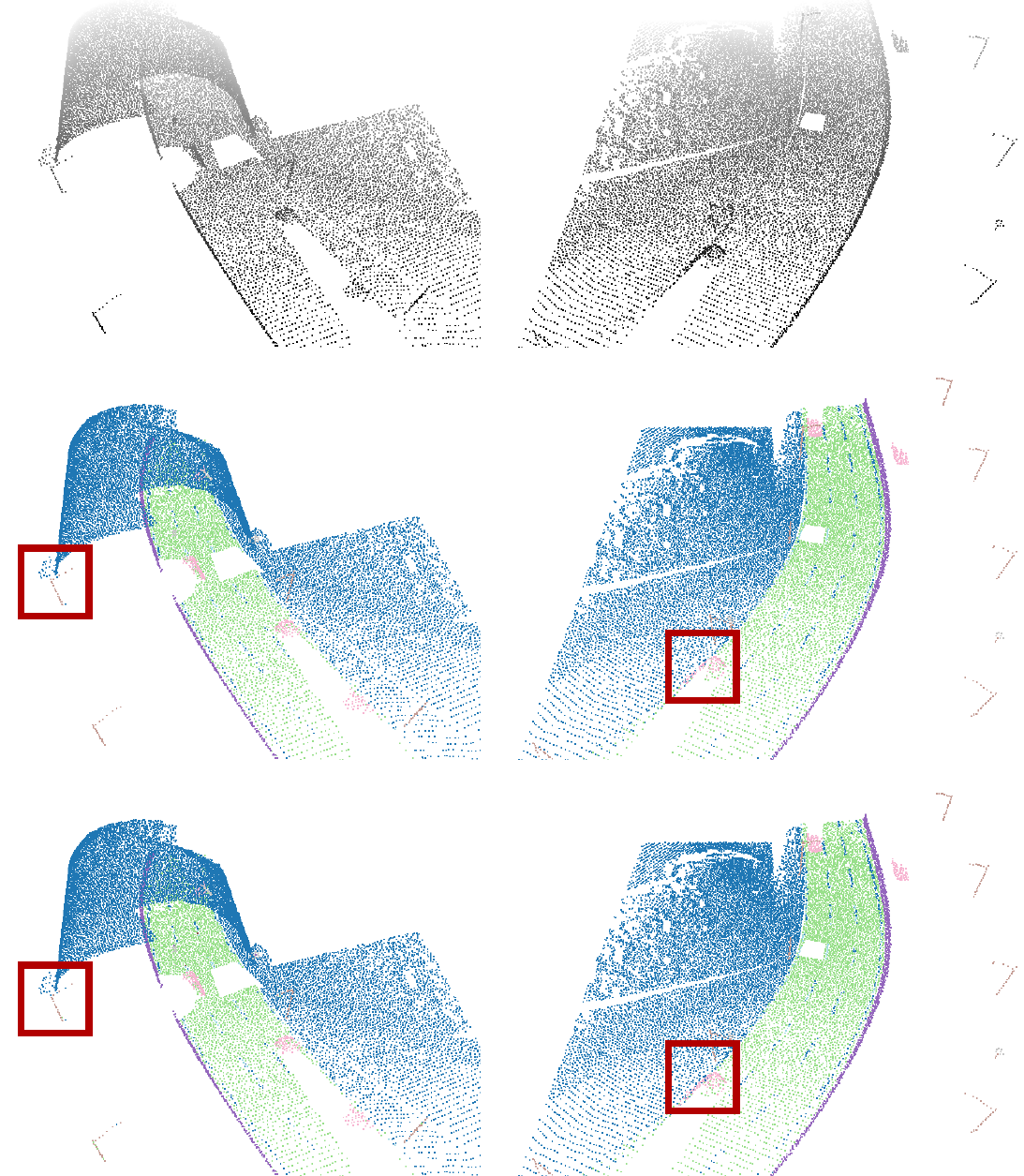}
\caption{\textbf{Visualization of 4D semantic segmentation.} Top: inputs. Middle: ground truth. Bottom: PoinTriE predictions. Red Box: Semantic details.}
\label{fig6}
\captionsetup{width=\textwidth}  
\end{wrapfigure}

\noindent additive adapter, also yields improved performance with our pretrained backbone. This result demonstrates the strong versatility of PoinTriE. When the fine-tuning paradigm is replaced with the proposed side tuning, performance is further enhanced. We then regulate the parameters of STS Net via gradient flow masking to fully exploit the model capacity. These observations also verify the overfitting phenomenon discussed in PointATA~\cite{sun2026alignadaptrethinkingparameterefficient}.

\textbf{Hyperparameters.} We conduct a pairwise analysis on the hyperparameters used during pretraining and fine-tuning, with results reported in Fig.~\ref{fig4}. In general, a larger number of rigid transformations consistently benefits task performance. However, an excessively large coefficient for motion consistency may introduce negative effects. Meanwhile, the masking ratio and LoRA rank should be carefully controlled to ensure the effectiveness of STS Net.

\textbf{Training details.} In Fig~\ref{fig5}, we summarize the pretraining data used by several 3D backbones~\cite{afham2022crosspoint,SUN2026112800,qi2023contrast} and report their corresponding performance. It can be observed that larger pretraining datasets consistently yield improved performance across different backbones. PoinTriE leverages the ShapeNet dataset and its projections, further augmenting its scale with text corpora and pseudo-trajectories to achieve efficient data exploitation. More importantly, our fine-tuning pipeline consumes only about one-third of the memory required by existing PEFT methods~\cite{zha2023instanceidpt,csa,sun2026alignadaptrethinkingparameterefficient}, making it practical for most research settings.

\section{Conclusion}
In this paper, we investigate the TriETL problem in point cloud video learning. Motivated by our insights into transfer learning, we formalize the TriETL co-design framework with theoretical proofs. Based on this formulation, we propose PoinTriE framework, which integrates the GMD Net and STS Net. Our method achieves superior performance with lower computational overhead across several point cloud video benchmarks. We note that to acquire rich supervisory signals, we have to fully exploit the original data and its variants, which results in longer pretraining time. Due to the complex analysis required to identify highly informative data, we leave this direction for future exploration.

\section*{Acknowledgements}
This work was supported in part by NSFC (No. 62125305) , the Natural Science Basis Research Plan in Shaanxi Province of China (No. 2025JC-JCQN-091) and Technology Innovation Leading Program of Shaanxi (Program No. 2024QY-SZX-23).

%
%
\bibliographystyle{splncs04}
\bibliography{main}
\end{document}